\documentclass{article}

\usepackage{PRIMEarxiv}

\usepackage[utf8]{inputenc}
\usepackage[T1]{fontenc}
\usepackage{mathptmx}                
\usepackage{amsmath,amssymb}
\usepackage{graphicx}
\usepackage{booktabs}
\usepackage{longtable}
\usepackage{array}
\usepackage{tabularx}
\usepackage{multirow}
\usepackage{xcolor}
\usepackage{float}
\usepackage{caption}
\usepackage{subcaption}
\usepackage{enumitem}
\usepackage{siunitx}
\usepackage{listings}
\usepackage{tikz}
\usepackage[siunitx, european resistors, european inductors, european voltages, european currents]{circuitikz}
\usepackage{url}

\usetikzlibrary{positioning,arrows.meta,calc,fit}

\tikzset{
  stage/.style={draw, rounded corners=1.2pt, minimum height=8.2mm,
                minimum width=17.5mm, align=center, font=\scriptsize,
                inner sep=1.5pt, fill=blue!6},
  aux/.style={draw, rounded corners=1.2pt, minimum height=6mm,
              minimum width=15mm, align=center, font=\scriptsize,
              inner sep=1.5pt, fill=gray!10},
  blk/.style={draw, minimum height=7mm, minimum width=11.5mm,
              align=center, font=\tiny, inner sep=1pt, fill=blue!6},
  hd/.style={draw, minimum height=4.6mm, minimum width=11.5mm,
             align=center, font=\tiny, inner sep=1pt, fill=orange!12},
  lbl/.style={font=\tiny, align=center},
  arr/.style={-{Stealth[length=1.6mm]}, line width=0.5pt},
  darr/.style={-{Stealth[length=1.6mm]}, line width=0.5pt, dashed},
}

\usepackage{amsthm}
\usetikzlibrary{shapes,arrows.meta,positioning,fit,calc}
\usepackage{algorithm}
\usepackage{algorithmic}

\usepackage[numbers,sort&compress]{natbib}

\usepackage{hyperref}
\hypersetup{colorlinks=true, linkcolor=blue!60!black, citecolor=blue!60!black, urlcolor=blue!70!black}

\title{Moving Like a Human: Ego-Motion-Normalized Temporal Signatures for Real-Time Aerial Person Tracking on Milliwatt-Class Hardware
}

\author{
  A. A. Jafari, C. Ozcinar\\
  University of Tartu \\
  Tartu, Estonia \\
  \texttt{\{akbar.anbar.jafari\}@ut.ee} \\
   \And
  G. Anbarjafari \\
  3S Holding OÜ \\
  Tartu, Estonia\\
  \texttt{shb@3sholding.com} \\
}

\begin{document}
\maketitle

\begin{abstract}
Follow-me person tracking must run on the drone itself, where affordable companion computers offer only a few effective int8 GFLOP/s. At typical follow distances a person spans $10$--$60$ pixels, indistinguishable from clutter and beyond the reach of single-frame appearance detectors. The missing evidence is \emph{temporal} and belongs in the \emph{input representation}, computed analytically, rather than in learned temporal machinery. EMTS-Det is a five-stage system that estimates ego-motion, converts each frame into ego-motion-normalized residual-motion channels, detects person centers with a 22k-parameter, 7.6-MFLOP network, tracks a locked target with a Kalman filter in stabilized coordinates, and verifies tracks with a 1-D convolutional classifier of human motion (ROC AUC $0.941$). Training uses a synthetic-motion curriculum with motion channels generated by the deployed ego-motion code. Multi-seed ablations locate the value in generalization: on held-out VisDrone-DET a luminance-only variant collapses to $0.051$ AP$_{25}$ versus $0.415$, as does YOLOv8n fine-tuned identically despite $1{,}100$ times the compute, while the deployed int8 detector reaches $0.694$ AP$_{25}$ in-domain and $0.444$ on this split. Temporal-shift modules lower accuracy, so the deployed detector is stateless. Silent int8 calibration failures are documented; min--max calibration with propagated caches matches float within $0.008$ AP. On a Raspberry Pi Zero~2W the pipeline runs at $31.85$~FPS with $0.462$ AP$_{25}$ and $0.714$ recall over $1{,}000$ real-world UAV videos, versus $1.95$~FPS and $0.172$ AP$_{25}$ for YOLOv8n. A $57$-second field sequence shows auto-lock at $1.3$\,s, $97.9\%$ lock recall, and recovery from all nine occlusions with zero false re-locks.
\end{abstract}


\section{Introduction}
Consumer and industrial drones increasingly ship a ``follow-me'' capability: the operator designates one person, and the aircraft keeps that person framed autonomously. The computer-vision problem hiding behind this feature is unusually hard on three axes simultaneously. First, the target is \emph{small}: at a conservative follow distance and altitude, a person occupies $10$--$60$ pixels of height in a $640{\times}512$ frame, well inside the regime where single-frame appearance is ambiguous~\cite{cheng2023towards}. Second, the camera \emph{moves}, and on a sub-250-gram airframe it moves violently --- wind gusts change attitude faster than any consumer gimbal fully compensates. Third, and decisively, the compute budget is \emph{milliwatt-class}: the companion computers that fit the cost and mass envelope of this product category, such as the Raspberry Pi Zero 2W (four Cortex-A53 cores at 1\,GHz, 512\,MB LPDDR2), sustain only about $2$--$4$ effective int8 GFLOP/s once memory bandwidth is accounted for.

The budget arithmetic rules out the standard answer. YOLOv8n~\cite{jocher2023yolov8}, the smallest member of the most widely deployed detector family, costs ${\approx}8.7$ GFLOPs at $640$-pixel input; dividing by the platform's effective throughput predicts the ${\sim}2$~FPS we indeed measure on-device. MobileNet-SSD~\cite{liu2016ssd,howard2017mobilenets} at ${\sim}1$~GFLOP reaches ${\sim}12$~FPS but sacrifices exactly the small-object accuracy the application needs. For $20$--$30$~FPS operation, \emph{the entire per-frame pipeline} --- capture, stabilization, detection, tracking, verification --- must fit in roughly $50$--$100$ MFLOPs. No single-frame detector of useful accuracy fits this envelope, and we contend none can: a $30{\times}60$-pixel gray smudge on scree simply does not contain enough appearance evidence to be classified reliably, no matter how cleverly capacity is spent.

What that smudge does have is a \emph{motion signature}. A walking person translates coherently over the ground while their silhouette deforms periodically at gait frequency; rocks do neither, swaying vegetation oscillates without translating, and drifting cloud shadows translate without articulation~\cite{johansson1973visual}. This information is temporal, spread over half a second of video, and extracting it is cheap \emph{if} the camera's own motion is first removed. That observation drives the design of this paper: rather than spending the FLOP budget on appearance capacity, we spend a few milliseconds of classical geometry to normalize away ego-motion, hand the network pre-computed motion evidence as input channels, and let a very small temporal network do the rest.

Concretely, we present EMTS-Det (Ego-Motion-normalized Temporal Signature Detection), a five-stage pipeline (Fig.~\ref{fig:pipeline}):

\begin{itemize}
\item \textbf{Stage A} estimates the inter-frame camera motion as a 4-DoF similarity transform from sparse Lucas--Kanade tracks with a RANSAC fit~\cite{lucas1981iterative,fischler1981ransac}, falling back to phase correlation~\cite{reddy1996fft} on texture-poor scenes.
\item \textbf{Stage B} warps the previous frame by that transform and produces three input channels --- luminance, ego-compensated residual-motion magnitude, and signed temporal difference --- so that everything bright in the motion channels is, by construction, \emph{independently moving}.
\item \textbf{Stage C} is a $21{,}941$-parameter anchor-free detector built from depthwise-separable blocks, costing $7.6$~MFLOPs per frame and emitting a center heatmap~\cite{zhou2019objects}, box geometry, and an 8-D re-identification embedding. The network is deliberately \emph{stateless}: controlled ablations (Sec.~\ref{sec:thesisablation}) show that once the input channels carry ego-normalized motion evidence, adding learned temporal machinery (Temporal Shift Modules~\cite{lin2019tsm}, in bidirectional or strictly-causal form) \emph{reduces} accuracy --- a negative result we quantify and act on.
\item \textbf{Stage D} tracks the single designated target with a constant-velocity Kalman filter~\cite{kalman1960} expressed in \emph{stabilized} coordinates, so gusts perturb the coordinate transform rather than the target's motion model, with two-key association (Mahalanobis gate and embedding similarity) and an explicit reacquisition mode.
\item \textbf{Stage E} verifies each track every eighth frame with an $8{,}289$-parameter 1-D CNN over a 16-frame sequence of ROI optical-flow descriptors --- a learned test of whether the tracked object \emph{moves like a human} --- and vetoes tracks that fail.
\end{itemize}

Two further contributions concern how such a system can be trained and deployed at all. Because public drone corpora such as VisDrone~\cite{zhu2021visdrone} provide dense person labels predominantly at a \emph{smaller} size operating point than long-range follow-me (Fig.~\ref{fig:sizehist}), we train on a \emph{synthetic-motion curriculum} that matches our deployment regime while still reporting generalization on VisDrone-DET. Critically, every synthetic clip is passed through the \emph{deployed} Stage A/B code to generate the motion channels, eliminating a train/deploy distribution gap by construction. On the deployment side we document two int8 calibration failures that produce silently wrong models~\cite{jacob2018quantization}: exponential-moving-average range estimation washes out the rare large activations of the box-size head and nearly halves AP even on our stateless detector ($0.361$ vs.\ $0.694$ AP$_{25}$), and stateful streaming variants additionally require calibration over \emph{propagated} cache states rather than zero-initialized ones --- demonstrated on our causal-TSM ablation model, where the combined repair recovers AP from $0.228$ to within $1\%$ of float. We believe the stateful failure mode in particular has not been documented.

On a held-out test set of $300$ composite clips over unseen background videos ($8{,}624$ ground-truth instances) the deployed int8 system reaches $0.694$ AP$_{25}$ / $0.434$ AP$_{50}$, within $0.008$ AP of its float reference. Controlled ablations (Tables~\ref{tab:thesisablation}, \ref{tab:archablation}) --- every variant retrained from scratch on the identical curriculum over three training seeds and evaluated in float under one protocol, with mean $\pm$ std reported --- isolate the thesis as a \emph{generalization} claim, and settle the architecture: adding temporal-shift modules to the detector \emph{lowers} accuracy in every configuration tested (bidirectional-then-converted and causal-from-scratch), while removing the motion channels barely moves the in-domain score but collapses held-out real VisDrone-DET performance $6\times$. Against single-frame detectors we report both the out-of-domain numbers (COCO YOLOv8n $0.160$, MobileNet-SSD $0.013$) and the fully symmetric one: YOLOv8n fine-tuned on the \emph{identical training mixture} --- Tier-1 composites, the same VisDrone train-split stills, and the same hard negatives --- reaches $0.886$ AP$_{25}$ on the bench at $8.7$ GFLOPs, yet only $0.103$ on held-out real VisDrone imagery, $4.3\times$ below the $1{,}100\times$-cheaper EMTS-Det ($0.444$). A classical compensated-differencing baseline reaches $0.120$ AP$_{25}$ at $5.2$ raw FP/frame. On $4{,}200$ person-free frames from unseen videos, the deployed detector emits $0.34$ raw FP/frame at the recall-oriented threshold and the Stage-E verifier halves that ($0.165$); the system-level stress-test KPI---verified false \emph{locks} per hour under autonomous auto-lock on person-free footage, a failure class absent in the operator-initiated product mode---is ${\sim}720$/hr (Sec.~\ref{sec:systemkpi}), with zero false re-locks during tracking on the field sequence. The tracker reacquires its target after forced occlusions in $93.6\%$ of $783$ trials. The complete pipeline executes in $3.3$~ms/frame measured end-to-end on an Apple-silicon laptop CPU restricted to the target's thread count; the per-stage budget analysis of Sec.~\ref{sec:latency} indicates comfortable real-time margins on Pi Zero~2W and similar low-compute devices, with on-device measurement left to future work. On a $57$-second field sequence, the system auto-locks at $t{=}1.3$\,s, maintains measured lock recall of $97.9\%$ on YOLO-confirmed frames, and suffers $88$ fewer detection dropouts than YOLOv8n (Figs.~\ref{fig:timeline}, \ref{fig:qualitative}).

\section{Related Work}

\subsection{Lightweight and Small-Object Detection}
The efficient-detector literature descends from SSD~\cite{liu2016ssd} and YOLO~\cite{redmon2016yolo} through depthwise-separable backbones~\cite{howard2017mobilenets,sandler2018mobilenetv2} to purpose-built mobile detectors such as PP-PicoDet~\cite{yu2021picodet} and YOLOv8n~\cite{jocher2023yolov8}. These models target the 1--10 GFLOP range --- one to two orders of magnitude above our budget --- and remain single-frame: their capacity is spent learning appearance invariances that our input representation supplies analytically. Anchor-free center-point heads~\cite{law2018cornernet,zhou2019objects,tian2019fcos} are particularly suited to tiny objects because they avoid anchor-matching pathologies at sub-cell scales; we adopt a CenterNet-style head~\cite{zhou2019objects} with focal heatmap supervision~\cite{lin2017focal}. Surveys of small-object detection~\cite{cheng2023towards} consistently identify insufficient per-frame evidence as the fundamental obstacle, which is precisely the deficiency temporal aggregation addresses.

\subsection{Efficient Temporal Modeling}
Temporal Segment Networks~\cite{wang2016tsn} aggregate frame-level features late; 3-D convolutions and their mobile derivatives~\cite{kondratyuk2021movinets} aggregate early but multiply cost. The Temporal Shift Module~\cite{lin2019tsm} obtains temporal receptive field at \emph{zero} FLOP overhead by shifting a fraction of channels across time before each 2-D convolution, and its uni-directional variant supports strictly causal streaming with cached channel slices. We evaluate TSM inside a detector rather than a classifier --- exporting the cached form with the caches as explicit graph inputs/outputs, which makes quantization calibration subtle (Sec.~\ref{sec:quant}) --- and reach a finding the efficient-video literature rarely tests for: when the \emph{input representation} already encodes ego-normalized motion, the learned shifts add cost without adding accuracy (Sec.~\ref{sec:thesisablation}), so the deployed detector omits them.

\subsection{Moving-Object Detection Under Camera Motion}
Detecting independent motion from a moving platform via background subtraction after global motion compensation is a classical line of work~\cite{yazdi2018moving}. These systems typically stop at ``something moved'': class-agnostic blobs, fragile to parallax and registration error. EMTS-Det instead treats the compensated residual as an \emph{input feature} to a learned detector, so registration noise becomes something the network learns to discount, and adds a downstream learned verifier that tests for specifically \emph{human} articulation patterns~\cite{johansson1973visual} rather than mere motion.

\subsection{Target-Locked Tracking}
Detection-based trackers~\cite{bewley2016sort,wojke2017deepsort,zhang2022bytetrack} associate per-frame detections to tracks with motion and appearance cues; FairMOT~\cite{zhang2021fairmot} showed detection and re-ID embeddings can share one network, which we adopt at drastically smaller scale (8-D embeddings from a 22k-parameter model). Our setting differs from multi-object tracking in that exactly one target matters after the operator's lock; this converts an open-world problem into re-detection of a known identity and motivates our two-key gate, template hygiene rules, and explicit reacquisition mode, evaluated with a dedicated occlusion protocol rather than MOT metrics~\cite{luiten2021hota}.

\subsection{Synthetic Training Data and Quantization}
Rendering people for training data is established~\cite{varol2017surreal}, as is domain randomization~\cite{tobin2017domain}. Our curriculum differs in \emph{what} is synthesized: not photorealistic appearance --- our input is grayscale plus motion, which has a far smaller sim-to-real gap than RGB --- but realistic \emph{motion}, both of the camera (replayed onto real plates) and of the person (articulated walk cycles, gait-phase limb warping). For deployment we use static int8 quantization~\cite{jacob2018quantization,krishnamoorthi2018quantizing,nagel2021white} with quantization-aware fine-tuning~\cite{esser2020lsq}; our contribution there is the identification and repair of a calibration failure specific to stateful streaming graphs.

\begin{figure}[!t]
\centering
\resizebox{\textwidth}{!}{%
\begin{tikzpicture}[node distance=4.5mm and 6.5mm]
  \node[aux] (cam) {Camera\\$640{\times}512$ @ 30\,Hz};
  \node[stage, right=of cam]  (A) {\textbf{A}\; Ego-motion\\sparse LK $+$ RANSAC\\$\to M_t$ (similarity)};
  \node[stage, right=of A]    (B) {\textbf{B}\; Motion channels\\warp, subtract\\$\to [L_t, R_t, D_t]$};
  \node[stage, right=of B]    (C) {\textbf{C}\; Micro-detector\\DWS backbone $+$ center head\\$\to$ dets $+$ 8-D emb.};
  \node[stage, right=of C]    (D) {\textbf{D}\; Track manager\\Kalman in stabilized frame\\two-key association};
  \node[stage, right=of D]    (E) {\textbf{E}\; Signature verifier\\1-D CNN on ROI flow\\$\to P(\mathrm{human})$};
  \node[aux, right=of E] (out) {Gimbal /\\flight cmd};
  \draw[arr] (cam) -- (A);
  \draw[arr] (A) -- (B) node[midway, above, lbl] {$M_t$};
  \draw[arr] (B) -- (C) node[midway, above, lbl] {$x_t$};
  \draw[arr] (C) -- (D) node[midway, above, lbl] {$\{d_i\}$};
  \draw[arr] (D) -- (E) node[midway, above, lbl] {ROI};
  \draw[arr] (E) -- (out);
  \draw[darr] (A.south) -- ++(0,-4.5mm) -| node[pos=0.25, below, lbl]
    {$M_t$ (coordinate stabilization)} (D.south);
  \draw[darr] (E.north) -- ++(0,3.6mm) -| node[pos=0.25, above, lbl]
    {verify / veto, gate template update} (D.north);
  \node[lbl, below=8.5mm of A.south] {every frame\\$4.5$\,ms$^\dagger$};
  \node[lbl, below=8.5mm of B.south] {every frame\\$2.6$\,ms$^\dagger$};
  \node[lbl, below=8.5mm of C.south] {every frame\\$0.66$\,ms$^\dagger$ (int8)};
  \node[lbl, below=8.5mm of D.south] {every frame\\$0.08$\,ms$^\dagger$};
  \node[lbl, below=8.5mm of E.south] {every 8th frame\\$0.35$\,ms$^\dagger$/call};
\end{tikzpicture}}
\caption{The EMTS-Det pipeline. Solid arrows carry per-frame data; dashed arrows carry the ego-motion transform into the tracker's coordinate stabilization and the verifier's veto back into track management. All heavy learning is concentrated in Stage~C, which the classical stages make small enough to afford: it receives ego-motion-normalized motion evidence instead of having to learn ego-motion invariance from data. $^\dagger$Measured on an Apple M-series CPU (Sec.~\ref{sec:latency}); the Pi-class budget analysis is in Table~\ref{tab:latency}.}
\label{fig:pipeline}
\end{figure}
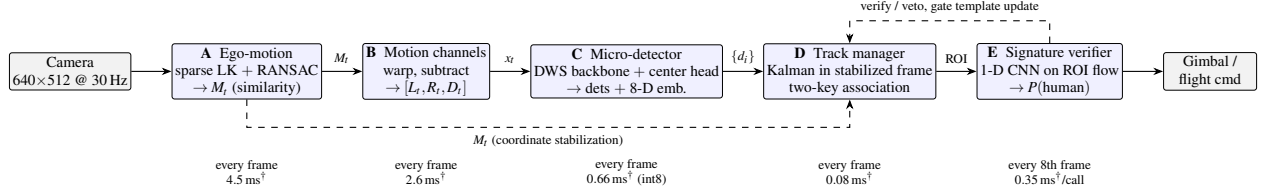

\section{The EMTS-Det Pipeline}
\label{sec:method}

\subsection{Problem Setting and Budget}
The system receives grayscale frames $I_t \in [0,1]^{H\times W}$ ($640{\times}512$) from a moving aerial camera at up to 30\,Hz and must output, after a one-tap lock at time $t_0$, the image-plane box of the designated person at every subsequent frame, tolerating occlusions, scale change, and camera shake, on a compute budget of ${\sim}100$ MFLOPs/frame end-to-end. We denote by $M_t \in \mathbb{R}^{2\times3}$ the affine matrix mapping coordinates of frame $t{-}1$ into frame $t$.

\subsection{Stage A: Ego-Motion Estimation}
\label{sec:stageA}
On a $320{\times}256$ downsample, up to $120$ Shi--Tomasi corners~\cite{shi1994good} are tracked from $I_{t-1}$ to $I_t$ with pyramidal Lucas--Kanade~\cite{lucas1981iterative}. From the surviving correspondences $\{(\mathbf{p}_i, \mathbf{q}_i)\}$ a 4-DoF similarity transform
\begin{equation}
M_t = \begin{bmatrix} s\cos\theta & -s\sin\theta & t_x \\ s\sin\theta & \phantom{-}s\cos\theta & t_y \end{bmatrix}
\label{eq:similarity}
\end{equation}
is estimated by RANSAC~\cite{fischler1981ransac} with a 2-pixel reprojection threshold, minimizing $\sum_i \rho\!\left(\lVert \mathbf{q}_i - M_t\tilde{\mathbf{p}}_i \rVert\right)$ over the inlier set. The 4-DoF model is deliberate: it absorbs the roll and zoom components of gust response that translation-only registration cannot, while remaining robust with far fewer correspondences than a full homography --- an appropriate trade at this image size, where the ground is locally near-planar and residual parallax is small relative to the $8$-pixel output stride of Stage C. If fewer than $15$ tracks survive or the inlier ratio falls below $0.4$ (uniform snow, water), the system falls back to Hanning-windowed phase correlation~\cite{reddy1996fft} on a $96{\times}72$ downsample, yielding a translation-only $M_t$. The estimate is used twice (Fig.~\ref{fig:pipeline}): to form the residual channels (Stage B) and to stabilize the tracker's coordinate frame (Stage D).

\subsection{Stage B: Ego-Motion-Normalized Input Channels}
\begin{figure}[!t]
\centering
\includegraphics[width=\columnwidth]{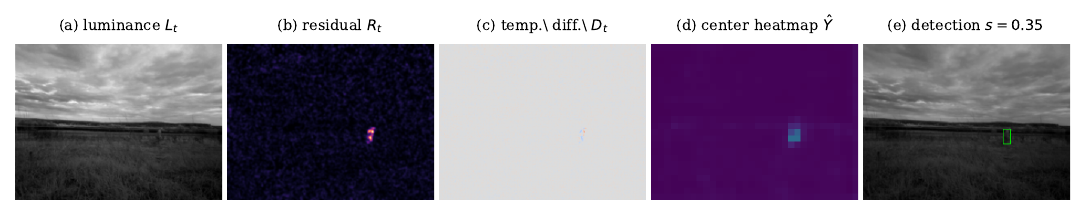}
\caption{The input representation on real drone footage (frame from the field sequence of Sec.~\ref{sec:qualitative}). (a)~Luminance: the person is a barely visible smudge left of center. (b)~Ego-compensated residual $R_t$: the person is the dominant response. (c)~Signed temporal difference $D_t$. (d)~Stage-C center heatmap $\hat{Y}$. (e)~Decoded detection of the deployed model (score inset) at $256{\times}192$ working resolution.}
\label{fig:channels}
\end{figure}

At the network's working resolution ($256{\times}192$), the previous frame is warped by $M_t$ and three channels are formed:
\begin{align}
\tilde{I}_{t-1} &= \mathcal{W}(I_{t-1};\, M_t), \qquad L_t = I_t, \nonumber\\
R_t &= G_{5\times5} * \big| I_t - \tilde{I}_{t-1} \big|, \qquad
D_t = I_t - \tilde{I}_{t-1},
\label{eq:channels}
\end{align}
where $\mathcal{W}$ denotes bilinear affine warping with border replication and $G_{5\times5}$ a Gaussian kernel. The detector input is $x_t = [L_t, R_t, D_t]$. Under exact registration of a static scene, $R_t$ and $D_t$ vanish everywhere except on independently moving objects; in practice registration error concentrates on high-gradient edges, a structured noise the detector learns to discount. Fig.~\ref{fig:channels} shows the three channels on real footage: a person invisible to the eye in $L_t$ is the single dominant blob in $R_t$.

Two properties of this representation carry the design. First, it \emph{bakes in} appearance invariance --- clothing color, one of the dominant nuisance factors in aerial person detection, never enters the network. Second, it shrinks the sim-to-real gap of synthetic training data (Sec.~\ref{sec:data}): motion channels computed by the same code on synthetic and real footage are far closer in distribution than rendered and real RGB.

\subsection{Stage C: Micro-Detector over Temporal Features}
\label{sec:stageC}

\subsubsection{Architecture}
\begin{figure}[!t]
\centering
\begin{tikzpicture}[node distance=2.6mm and 3.2mm]
  \node[blk, fill=gray!10] (in) {input\\$3{\times}192{\times}256$};
  \node[blk, right=of in]  (stem) {stem\\$3{\times}3$ s2\\$16{\times}96{\times}128$};
  \node[blk, right=of stem] (s1) {S1: 2$\times$DWS\\s2\\$24{\times}48{\times}64$};
  \node[blk, right=of s1]  (s2) {S2: 2$\times$DWS\\s2\\$40{\times}24{\times}32$};
  \node[blk, right=of s2]  (s3) {S3: 3$\times$DWS\\s2\\$64{\times}12{\times}16$};
  \node[blk, below=5.5mm of s2] (neck) {neck: $\uparrow$2 $+$ add\\dw-smooth\\$40{\times}24{\times}32$};
  \node[hd, below=5.5mm of neck, xshift=-19.5mm] (hm) {$\hat{Y}$ heatmap\\$1{\times}24{\times}32$};
  \node[hd, right=1.8mm of hm] (wh) {$\hat{S}$ size\\$2{\times}24{\times}32$};
  \node[hd, right=1.8mm of wh] (off) {$\hat{O}$ offset\\$2{\times}24{\times}32$};
  \node[hd, right=1.8mm of off] (emb) {$\hat{E}$ embed\\$8{\times}24{\times}32$};
  \draw[arr] (in) -- (stem);
  \draw[arr] (stem) -- (s1);
  \draw[arr] (s1) -- (s2);
  \draw[arr] (s2) -- (s3);
  \draw[arr] (s3.south) |- node[pos=0.75, above, lbl] {$1{\times}1$ lat.} (neck.east);
  \draw[arr] (s2.south) -- node[midway, right, lbl] {skip} (neck.north);
  \draw[arr] (neck.south) -- ++(0,-2.2mm) -| (hm.north);
  \draw[arr] (neck.south) -- ++(0,-2.2mm) -| (wh.north);
  \draw[arr] (neck.south) -- ++(0,-2.2mm) -| (off.north);
  \draw[arr] (neck.south) -- ++(0,-2.2mm) -| (emb.north);
\end{tikzpicture}
\caption{Stage-C architecture (deployed, stateless form). All four heads are single $1{\times}1$ convolutions on the shared neck feature. In the TSM ablation variants, a temporal shift (\ref{eq:tsm})/(\ref{eq:causaltsm}) precedes each DWS (depthwise-separable) block's depthwise convolution at zero FLOP cost. Total: $21{,}941$ parameters, $7.6$ MFLOPs/frame.}
\label{fig:arch}
\end{figure}
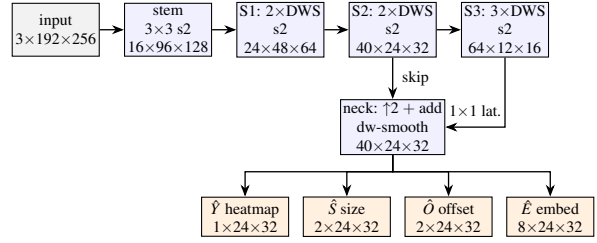

The detector (Fig.~\ref{fig:arch}, Table~\ref{tab:arch}) is a depthwise-separable~\cite{howard2017mobilenets} encoder of three stages with a single lateral top-down fusion, emitting predictions at output stride $8$ ($32{\times}24$ grid). The deployed network is \emph{stateless}: each frame's ego-normalized channel stack $x_t$ is processed independently, and all temporal evidence enters through the input representation of Stage~B. This is a conclusion, not merely a choice. The natural alternative --- and our original design --- inserts a Temporal Shift Module~\cite{lin2019tsm} before each depthwise convolution: with channel tensor $z_t \in \mathbb{R}^{C\times h\times w}$ and fold size $f = \lfloor C/8 \rfloor$, offline training uses the bidirectional shift
\begin{equation}
\mathrm{TSM}(z)_t = \left[\, z_{t-1}^{(1:f)};\; z_{t+1}^{(f+1:2f)};\; z_t^{(2f+1:C)} \,\right],
\label{eq:tsm}
\end{equation}
mixing information across three time steps at zero FLOP and zero parameter cost, while streaming inference replaces it with a \emph{causal} cached form that takes both shifted folds from the previous frame's slices,
\begin{equation}
\mathrm{TSM}^{\mathrm{c}}(z)_t = \left[\, c_{t-1};\; z_t^{(2f+1:C)} \,\right], \qquad
c_t \leftarrow z_t^{(1:2f)},
\label{eq:causaltsm}
\end{equation}
with the seven caches $c^{(1)}\ldots c^{(7)}$ ($18{,}816$ values) exported as explicit graph inputs and outputs, making that variant a pure function
\begin{equation}
\big(\hat{Y}, \hat{S}, \hat{O}, \hat{E},\, c'^{(1..7)}\big) = F_\theta\big(x_t,\, c^{(1..7)}\big).
\label{eq:statefn}
\end{equation}
Controlled ablations at identical training budget (Sec.~\ref{sec:thesisablation}) show that every TSM configuration --- bidirectional-then-converted per (\ref{eq:tsm})/(\ref{eq:causaltsm}), and a variant trained with the causal shift semantics from the first gradient step --- \emph{underperforms} the plain stateless network on both the in-domain bench and real held-out imagery. The analytic motion channels already deliver the short-horizon temporal evidence the shifts were meant to aggregate, so the learned machinery pays its costs (an eighth of the channels displaced per block, plus quantization complexity) without adding information. We therefore deploy the stateless form and retain the TSM variant as a controlled subject for the stateful-quantization study of Sec.~\ref{sec:quant}.

The heads follow CenterNet~\cite{zhou2019objects}: a person-center heatmap $\hat{Y} \in [0,1]^{24\times32}$, log-scale size $\hat{S} \in \mathbb{R}^{2\times24\times32}$, sub-cell offset $\hat{O} \in \mathbb{R}^{2\times24\times32}$, and an 8-D embedding field $\hat{E}$ trained for re-identification (Sec.~\ref{sec:stageD}). Detections are decoded from local maxima of $\hat{Y}$ (3$\times$3 max-pool non-maximum suppression), with box $(w,h) = 8\exp(\hat{S})$ and center $(u + \hat{O}_u, v + \hat{O}_v)\cdot 8$. At $30$--$60$~px person height, a target covers 1--4 grid cells --- sufficient for center-point detection, with the offset head recovering sub-cell precision.

The full network has $21{,}941$ parameters and costs $3.82$ MMACs ($7.6$ MFLOPs) per frame --- $1{,}100\times$ less than YOLOv8n --- and $73$\,KiB as an int8 flatbuffer. Anchor-free decoding was chosen over anchor matching specifically for this scale regime: at $10$--$20$-px object sizes, anchor IoU assignment is dominated by quantization noise, while center-point assignment degrades smoothly.

\begin{table}[!t]
\caption{Stage-C Compute Breakdown (Per Frame, $3{\times}192{\times}256$ Input)}
\label{tab:arch}
\centering
\begin{tabular}{lccr}
\toprule
Stage & Output shape & Params & MMACs \\
\midrule
Stem ($3{\times}3$, s2)      & $16{\times}96{\times}128$ & 464    & 1.33 \\
S1 (2$\times$ DWS)           & $24{\times}48{\times}64$  & 1\,496 & 1.01 \\
S2 (2$\times$ DWS)           & $40{\times}24{\times}32$  & 3\,424 & 0.60 \\
S3 (3$\times$ DWS)           & $64{\times}12{\times}16$  & 12\,984 & 0.59 \\
Neck (lat.\ $+$ dw-smooth)   & $40{\times}24{\times}32$  & 3\,120 & 0.19 \\
Heads ($4\times$ $1{\times}1$) & $13{\times}24{\times}32$ & 533   & 0.10 \\
\midrule
\textbf{Total} & & \textbf{21\,941} & \textbf{3.82} \\
\bottomrule
\end{tabular}
\end{table}

\subsubsection{Losses}
Training clips provide per-frame ground-truth boxes. Centers are splatted onto the heatmap target $Y$ with size-adaptive Gaussians as in~\cite{zhou2019objects}, and the heatmap is supervised with the penalty-reduced focal loss~\cite{lin2017focal,law2018cornernet}
\begin{equation}
\mathcal{L}_{\mathrm{hm}} = \frac{-1}{N}\!\sum_{uv}\!
\begin{cases}
(1{-}\hat{Y}_{uv})^{2}\log \hat{Y}_{uv} & \! Y_{uv}{=}1,\\[2pt]
(1{-}Y_{uv})^{4}\,\hat{Y}_{uv}^{2}\log(1{-}\hat{Y}_{uv}) & \!\text{else},
\end{cases}
\label{eq:focal}
\end{equation}
with $N$ the number of positives. Size and offset use masked $L_1$ losses $\mathcal{L}_{\mathrm{wh}}, \mathcal{L}_{\mathrm{off}}$ at positive cells. The embedding field is trained with a supervised contrastive loss~\cite{khosla2020supcon} over the set $\mathcal{B}$ of embeddings gathered at ground-truth centers across the batch and across time, with identity labels $y_i$ (an identity persists across the frames of a clip):
\begin{equation}
\mathcal{L}_{\mathrm{emb}} = \sum_{i\in\mathcal{B}} \frac{-1}{|P(i)|} \sum_{p\in P(i)} \log
\frac{\exp(\mathbf{e}_i^{\!\top}\mathbf{e}_p/\tau)}{\sum_{a \neq i}\exp(\mathbf{e}_i^{\!\top}\mathbf{e}_a/\tau)},
\label{eq:supcon}
\end{equation}
where $P(i) = \{p : y_p = y_i, p \neq i\}$ and $\tau{=}0.1$. This teaches the embedding to be stable for the \emph{same} person across frames --- including across the scale sweeps present in the training data (Sec.~\ref{sec:data}) --- while separating distinct people, which is what powers target lock-on. The total objective is
\begin{equation}
\mathcal{L} = \mathcal{L}_{\mathrm{hm}} + 0.1\,\mathcal{L}_{\mathrm{wh}} + \mathcal{L}_{\mathrm{off}} + 0.5\,\mathcal{L}_{\mathrm{emb}}.
\label{eq:total}
\end{equation}

\subsection{Stage D: Target-Locked Tracking in Stabilized Coordinates}
\label{sec:stageD}
Let $G_t = H(M_t)\,G_{t-1}$ (with $H(\cdot)$ the $3{\times}3$ homogeneous lift, $G_0 = I$) accumulate ego-motion since lock time. All track state lives in the \emph{stabilized} frame: a detection at image position $\mathbf{u}$ is first mapped to $\mathbf{u}^{s} = G_t^{-1}\tilde{\mathbf{u}}$. The consequence is architectural, not cosmetic: a wind gust changes $G_t$, not the target's state, so the constant-velocity model below remains a good model of a \emph{person} even when the camera is anything but constant-velocity.

The track carries state $\mathbf{x} = [x, y, \dot{x}, \dot{y}, w, h]^\top$ under a constant-velocity Kalman filter~\cite{kalman1960} with measurement $\mathbf{z} = [x, y, w, h]^\top$; predict/update follow the standard equations with process noise $Q = \mathrm{diag}(q, q, 4q, 4q, \tfrac{q}{4}, \tfrac{q}{4})$, $q{=}1$, and measurement noise $R = \mathrm{diag}(4, 4, 9, 9)$ px$^2$. At lock-on, the operator's tap selects detection $d^\ast$; its embedding initializes the target template $\mathbf{m} \leftarrow \mathbf{e}^\ast/\lVert\mathbf{e}^\ast\rVert$ and its height the reference scale $h_{\mathrm{lock}}$.

\subsubsection{Two-Key Association}
While locked, a detection $d_i$ is admissible if it passes \emph{both} gates:
\begin{equation}
\gamma_i^2 = \boldsymbol{\nu}_i^\top S^{-1} \boldsymbol{\nu}_i < 9.21, \qquad
s_i = \mathbf{m}^\top \mathbf{e}_i > \tau_e(h_i),
\label{eq:gates}
\end{equation}
where $\boldsymbol{\nu}_i$ is the positional innovation, $S$ its covariance ($\chi^2_{0.99}$ gate with 2 DoF), and among admissible candidates the one minimizing $\gamma_i^2 - 4 s_i$ is selected. The embedding gate is \emph{scale-aware}:
\begin{equation}
\tau_e(h) = \begin{cases} \tau_0, & h \le 1.5\,h_{\mathrm{lock}}, \\
\max\!\big(0.2,\ 1.5\,\tau_0\, h_{\mathrm{lock}}/h\big), & h > 1.5\,h_{\mathrm{lock}},
\end{cases}
\label{eq:scalegate}
\end{equation}
with $\tau_0{=}0.5$. The rationale: embeddings from a network trained predominantly on small targets go out of distribution when the target closes range and grows; (\ref{eq:scalegate}) transfers veto power from appearance to the (scale-independent) spatial gate exactly when appearance becomes unreliable. $h_{\mathrm{lock}}$ itself is tracked with a slow EMA so the reference adapts over minutes, not frames.

\subsubsection{Template Hygiene}
The template is updated at two rates. On every accepted association a slow blend $\mathbf{m} \leftarrow \mathrm{norm}(0.98\,\mathbf{m} + 0.02\,\mathbf{e})$ tracks gradual appearance change (a person growing in the frame can then never drift far from the template); a faster update ($\tau{=}0.95$) is applied only on frames the Stage-E verifier has approved, so that sustained template adaptation is conditioned on the track actually moving like a human. This two-rate rule is what prevents the classic template-drift failure in which a tracker slowly transfers its lock onto a false positive.

\subsubsection{Reacquisition}
After $8$ consecutive gate failures, or a verifier veto, the track enters reacquisition: the filter coasts on its prediction and every incoming detection is scored against the \emph{frozen} template, re-locking on $s_i > 0.65$. A second, spatial path handles the out-of-distribution case that motivated (\ref{eq:scalegate}): if a detection falls inside the coasted Mahalanobis gate ($\gamma^2 < 16$) for three \emph{consecutive} frames, motion consistency overrides appearance --- the track re-locks on it and rebuilds the template from its embedding. Requiring temporal persistence keeps this fallback from being hijacked by transient clutter, since uncorrelated false positives rarely persist inside a moving gate.

\subsection{Stage E: Temporal-Signature Verification}
\label{sec:stageE}
The detector is deliberately tuned for recall; the verifier restores precision by testing the one property clutter cannot fake: human articulation. Every frame, the track's ROI in the Stage-B pair $(\tilde{I}_{t-1}, I_t)$ is resampled to $24{\times}48$ and dense Farneb\"ack flow~\cite{farneback2003two} is computed --- this flow is \emph{already ego-compensated} because the first argument is the warped previous frame. The ROI is divided into a $2{\times}4$ grid (head to legs); each cell contributes its mean flow magnitude and a magnitude-weighted 4-bin orientation histogram, giving a $2\cdot4\cdot(1{+}4) = 40$-D descriptor per frame. A 16-frame window ($0.53$\,s at 30\,Hz --- roughly one gait cycle) forms the input $\Phi \in \mathbb{R}^{40\times16}$ to an $8{,}289$-parameter 1-D CNN (two Conv1d layers, global pooling, two linear layers; $0.12$ MMACs/call) emitting $P(\mathrm{human\ motion})$.

The head-to-legs grid gives the network the spatial layout needed to discover the discriminative pattern: leg-region flow oscillating at gait frequency (walking $1.5$--$2.5$\,Hz falls well inside the window's Nyquist range at 30\,Hz sampling) superimposed on coherent whole-body translation. Its learned rejections are exactly the aerial false-positive taxonomy: vegetation oscillates without translating; shadows translate without articulating; registration noise does neither coherently. Verification runs every $8$th frame per track and feeds back into Stage D (veto and template gating), so its amortized cost is under $0.05$\,ms/frame. Because detector and verifier fail for \emph{nearly independent} reasons --- one looks at a snapshot of motion evidence, the other at half a second of its temporal structure --- the system's false-positive rate approaches the product of the two individual rates at matched recall.

\subsection{Multi-Rate Schedule}
The stages run at three rates: A--D every frame, E every eighth frame per track, and the template fast-update only on verified frames. Everything except Stage C is NumPy/OpenCV; Stage C and E are int8 TFLite executed by LiteRT/XNNPACK~\cite{litert} on 4 threads. The measured cost structure (Fig.~\ref{fig:pipeline}, Table~\ref{tab:latency}) confirms the design premise: the \emph{learned} components consume under $10\%$ of the per-frame budget, because the classical stages made them small enough.

\section{Synthetic Motion Curriculum}
\label{sec:data}
No public dataset supplies dense person-labeled drone video at our \emph{operating point} (10--60-px targets at working resolution, aggressive ego-motion, person-free hard negatives); Fig.~\ref{fig:sizehist} quantifies the mismatch with VisDrone-DET~\cite{zhu2021visdrone}, where $96\%$ of person instances fall below $12$\,px. Single-frame datasets cannot supervise temporal machinery at all. We therefore construct training data whose \emph{motion} is realistic even where its appearance is synthetic, in three tiers. Throughout, one rule is absolute: every clip --- synthetic or real --- is passed through the \emph{deployed} Stage A/B implementation to produce the input channels. The network never sees an analytically synthesized motion channel, so there is no train/deploy representation gap to bridge.

\subsubsection{Tier 1: Rendered Actors on Real Plates}
Articulated human characters with walk-cycle animations are rendered in Blender~\cite{blender} under randomized camera pitch, orbit, distance, and lighting, producing RGBA sprite sequences. Each training clip composites $1$--$3$ actors onto an $18$-frame window of \emph{real} drone footage (so Stage A sees genuine ego-motion), at log-uniform heights spanning $24$--$220$ canvas pixels ($9$--$83$\,px at working resolution) --- tiny distant targets remain the common case, but close-range scales are represented --- with a per-frame scale drift of up to $\pm2.5\%$ so the supervised-contrastive loss (\ref{eq:supcon}) sees the \emph{same identity across a scale sweep}, which is what later lets the tracker hold lock through an approach. Luminance is harmonized to the local plate statistics, and edges are alpha-feathered. Animal actors (a quadruped walk cycle) are composited \emph{unlabeled}: they are moving objects the detector must not fire on and the verifier must reject --- a moving hard negative. $400$ training and $300$ test clips are generated, with train and test drawing on disjoint background videos.

\subsubsection{Tier 2: Pseudo-Video from Static Aerial Imagery}
Static drone imagery with person boxes (VisDrone~\cite{zhu2021visdrone} and a camouflage-person collection) is converted to 4-frame pseudo-video on the fly during pretraining: the whole image follows a smooth random similarity trajectory (drone motion, up to $4$\,px/frame drift, $0.008$\,rad/frame roll, $0.6\%$ scale rate), while each person crop follows its \emph{own} trajectory with an oscillating lower-body shear as a crude gait proxy, plus sensor noise. A zoom augmentation re-crops the scene around a person to a randomized target height (log-uniform $28$--$130$ canvas px), covering the scale range including close-range targets. This tier is unlimited in volume, costs nothing to store, and teaches precisely the residual-motion prior: \emph{a thing that moves differently from the background}.

\subsubsection{Hard Negatives}
Person-free clips are mined from real drone footage of the deployment environment (verified person-free by an oversensitive HOG screen~\cite{dalal2005hog} plus the person-free recording protocol), capturing swaying vegetation, water shimmer, moving shadows, and registration noise. $640$ such clips participate in training, and $150$ clips from held-out videos form the false-positive test bench. Hard negatives are a first-class dataset here, not an afterthought: they are disproportionately responsible for the final FP rate.

\subsubsection{Training Schedule}
Training proceeds in three phases on a laptop GPU (Apple MPS), total wall-clock under $15$ minutes at this model size. \emph{Phase 1}: 4 epochs $\times$ 2000 Tier-2 clips (AdamW, lr $1.5{\times}10^{-3}$, cosine decay, batch 8 clips $\times$ 4 frames). \emph{Phase 2}: 14 epochs on a $0.5/0.3/0.2$ mixture of Tier-1 / hard negatives / Tier-2, with negatives thereby present in every batch. \emph{Phase 3}: 4 epochs of quantization-aware fine-tuning with per-channel symmetric weight fake-quantization and moving-average activation observers~\cite{jacob2018quantization,esser2020lsq}. The verifier trains separately (30 epochs, $<$5\,s) on descriptor sequences extracted from Tier-1 ground-truth tracks (positives), hard-negative clips (negatives), and --- importantly --- from \emph{detector-proposed} boxes on both, so it is robust to the sloppy, jittering boxes it will actually receive at inference.

\section{Integer Deployment: Calibration Failure Modes}
\label{sec:quant}
The deployed detector is exported to TFLite and statically quantized to int8 (per-channel symmetric weights, per-tensor asymmetric activations~\cite{jacob2018quantization,krishnamoorthi2018quantizing}) after quantization-aware fine-tuning. Deployment surfaced calibration failure modes that produce silently wrong models rather than errors; we document them as general guidance --- one applies to any detector with regression heads, the others to any \emph{stateful} streaming network, which our TSM ablation variant (\ref{eq:statefn}) lets us demonstrate under controlled conditions.

\subsubsection{Min--Max vs.\ Moving-Average Range Estimation}
\label{sec:minmax}
Standard calibrators smooth per-batch activation ranges with an exponential moving average (smoothing factor $0.95$), which is appropriate for classifiers where ranges are stationary across samples. It is \emph{wrong} for regression heads with heavy-tailed outputs: the log-size head $\hat{S}$ emits rare large values --- precisely the large close-range boxes --- that the moving average washes out, silently clipping every big box the network predicts. On the deployed (stateless) model, moving-average calibration costs nearly half the accuracy ($0.361$ vs.\ $0.694$ AP$_{25}$ on the full bench); true min--max estimation over the calibration stream brings int8 to within $0.008$ AP$_{25}$ of the float reference (Table~\ref{tab:detection}). Note this failure is \emph{not} an artifact of statefulness --- it will affect any quantized detector whose size head has heavy tails.

\subsubsection{Stateful Graphs: Cache Calibration Requires Propagated State}
Streaming networks that carry temporal state as explicit graph I/O --- our causal-TSM variant (\ref{eq:statefn}), and more broadly any cached-attention or recurrent detector --- add a second trap. The caches are \emph{inputs}, so calibration must estimate their ranges too; feeding zero-initialized caches (the na\"ive choice) calibrates them on a distribution the network sees only at stream start, clipping all steady-state temporal information to zero. Our calibrator instead \emph{streams} each calibration clip through the float model, snapshotting the true $(x_t, c^{(1..7)})$ pairs at every step ($100$ clips of $10$ consecutive frames), so cache ranges reflect steady-state statistics. On the TSM variant the two repairs together recover AP$_{25}$ from $0.228$ to $0.543$ against a float reference of $0.548$ (Table~\ref{tab:quantablation}). A related pitfall: an in-graph $L_2$-normalization of the embedding head divides by a norm that quantizes to zero on an all-zero input, crashing integer inference; we remove the normalization from the graph and renormalize in the (floating-point) decoder. We expect all three lessons --- min--max ranges for heavy-tailed regression heads, propagated-state calibration, and no in-graph normalization of quantized embeddings --- to transfer to any small detector deployed at int8.

\section{Experiments}
\label{sec:experiments}

\subsection{Setup}
\label{sec:setup}
\textbf{Data splits.} All evaluations use background videos never seen in training. The detection test set is $300$ Tier-1 composite clips ($18$ frames each; evaluation on frames $5$--$18$ after temporal-state warm-up) over held-out videos, containing $8{,}624$ ground-truth person instances with the size distribution of Fig.~\ref{fig:recallheight}. The false-positive bench is $300$ person-free clips ($4{,}200$ evaluated frames) from three held-out videos of distinct environments. We additionally evaluate on the held-out $15\%$ image split of VisDrone-DET~\cite{zhu2021visdrone} person annotations ($851$ images never used by Tier-2 pretraining, converted to 4-frame pseudo-clips with the deployed Stage~A/B code) and report qualitative results on an independently captured field sequence (Sec.~\ref{sec:qualitative}). Finally, Sec.~\ref{sec:pizero} adds an on-device, in-the-wild evaluation on $1{,}000$ UAV videos collected from the Ukrainian Drone Force Telegram channel, a corpus disjoint from all training data.

\textbf{Metrics.} AP at IoU $0.25$ and $0.5$ --- we report AP$_{25}$ as the headline localization metric and justify AP$_{50}$ with a center-error analysis (Sec.~\ref{sec:localization}). At $10$--$20$-px object height, a $2$-px registration offset moves IoU across the $0.5$ boundary, so AP$_{50}$ measures annotation-scale noise as much as detection quality~\cite{cheng2023towards}; the application requires a lock point, not a tight box. False positives per frame on person-free footage, swept over the score threshold, with and without Stage-E gating (verifier threshold $0.5$). \emph{System KPI}: verified false \emph{locks} per hour on person-free footage under the auto-lock protocol (Sec.~\ref{sec:systemkpi}). Field lock recall and center error against YOLO pseudo-labels on the close/mid-range portion of the field sequence. Verifier ROC AUC on a held-out $15\%$ split of its dataset. Reacquisition success under a forced-occlusion protocol. Latency per stage.

\textbf{Hardware.} Timings for the controlled bench experiments are on an Apple M-series laptop CPU restricted to 4 threads (matching the core count of Pi-class targets), int8 models under LiteRT/XNNPACK~\cite{litert}. Sec.~\ref{sec:latency} gives a per-stage budget analysis for Pi Zero~2W and similar low-compute devices, and Sec.~\ref{sec:pizero} converts that budget into a measurement: the full pipeline and all baselines are evaluated end-to-end on a Raspberry Pi Zero~2W over the $1{,}000$-video real-world corpus (Table~\ref{tab:ukrainian_droneforce_pi_zero}); the released \texttt{scripts/measure\_pi.sh} reproduces the on-device measurement as a one-command exercise.

\subsection{Detection Accuracy and Quantization Fidelity}
\begin{figure}[!t]
\centering
\includegraphics[width=\columnwidth]{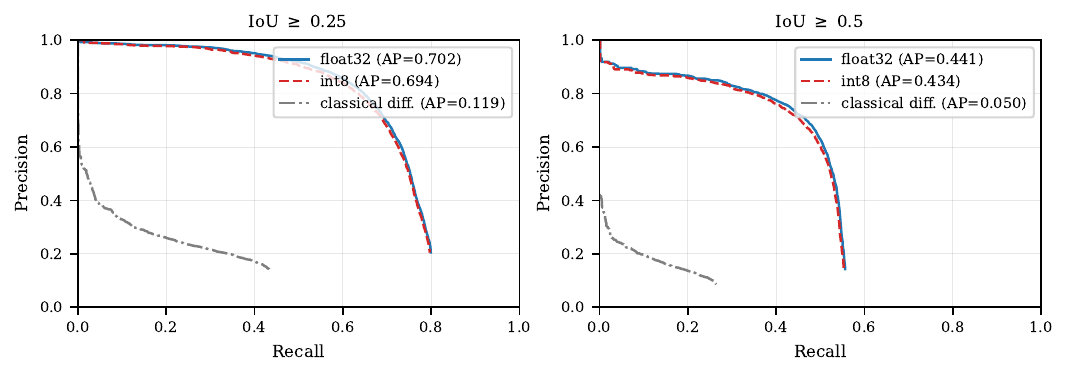}
\caption{Precision--recall of the deployed stateless model on the held-out bench at IoU $\ge 0.25$ (left) and $\ge 0.5$ (right), with the classical compensated-differencing baseline (dash-dot) for reference. The int8 model (dashed) tracks float within $0.008$ AP; the AP$_{50}$ drop relative to AP$_{25}$ is concentrated at sub-$20$-px targets, where a $2$-px extent error crosses the IoU $0.5$ boundary (Sec.~\ref{sec:localization}).}
\label{fig:pr}
\end{figure}

\begin{figure}[!t]
\centering
\includegraphics[width=0.42\columnwidth]{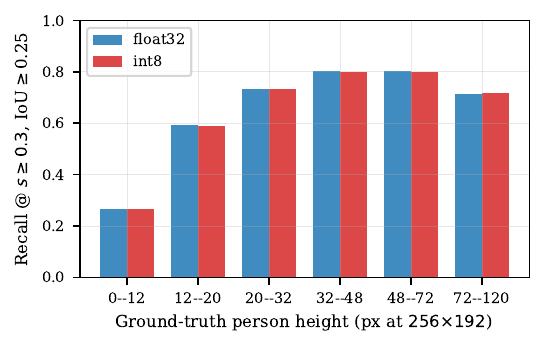}
\caption{Recall at the operating point ($\tau_s{=}0.3$, IoU $\ge 0.25$) by ground-truth height. Performance is strong throughout the follow-me range ($\ge$32\,px) and degrades gracefully toward the $\le$12-px extreme, where a person spans a single output cell.}
\label{fig:recallheight}
\end{figure}

Table~\ref{tab:detection} and Fig.~\ref{fig:pr} present precision--recall behavior of the deployed int8 model. Two observations. First, absolute AP$_{25}$ of $0.694$ on this bench must be read against the bench's difficulty: half the ground-truth instances are under $28$\,px tall (Fig.~\ref{fig:recallheight}), the clips contain real ego-motion, and $10\%$ of composited actors are unlabeled animals that count as false positives if detected as persons. Recall at the operating threshold reaches $0.80$ across the $32$--$72$\,px band --- the size range of the actual follow-me scenario --- and degrades gracefully, not catastrophically, below it ($0.73$ at $20$--$32$\,px, $0.59$ at $12$--$20$\,px). Second, quantization is essentially free \emph{only} once calibrated correctly: with the min--max recipe of Sec.~\ref{sec:minmax}, int8 tracks the float reference within $0.008$ AP$_{25}$ ($0.694$ vs.\ $0.702$; Fig.~\ref{fig:pr}), whereas the library-default moving-average calibration silently halves accuracy on the identical weights ($0.361$). The stateful-graph calibration study of Table~\ref{tab:quantablation} uses the TSM variant on a held-out $120$-clip subset.

\begin{table}[!t]
\caption{Detection Accuracy on Held-Out Composite Clips ($8{,}624$ GT, Deployed Model)}
\label{tab:detection}
\centering
\begin{tabular}{lccc}
\toprule
Model & AP$_{25}$ & AP$_{50}$ & Size \\
\midrule
EMTS-Det float32 (reference) & 0.702 & 0.441 & 112\,KiB \\
EMTS-Det int8 (deployed, min--max calib.)  & 0.694 & 0.434 & 73\,KiB \\
EMTS-Det int8 (moving-average calib.) & 0.361 & 0.063 & 73\,KiB \\
\bottomrule
\end{tabular}
\end{table}

\subsection{Ablation: Calibrating a Stateful Streaming Graph}
The deployed detector is stateless, but the failure modes of Sec.~\ref{sec:quant} matter most for the growing class of \emph{stateful} streaming networks, and our causal-TSM ablation variant provides a controlled subject. Table~\ref{tab:quantablation} isolates the calibration contribution on that variant's identical float weights. The standard recipe --- moving-average ranges over zero-initialized caches --- loses $58\%$ of AP$_{25}$ relative to float; more calibration data does not help (rows 1--2), because the failure is the estimator, not the sample size: each additional batch \emph{pulls the range estimate back toward the common case}, clipping the heavy tail of the size head harder. True min--max estimation over streamed, propagated cache states recovers float accuracy. Together with the stateless-model result of Table~\ref{tab:detection} (moving-average costs $0.33$ AP$_{25}$ even with no caches to calibrate), this separates the two failure mechanisms and gives practitioners an actionable recipe for deploying either class of network at int8.

\begin{table}[!t]
\caption{Int8 Calibration Ablation on the Stateful (Causal-TSM) Variant (Same Float Weights, AP on $120$-Clip Held-Out Subset, $3{,}528$ GT)}
\label{tab:quantablation}
\centering
\begin{tabular}{lcc}
\toprule
Calibration & AP$_{25}$ & AP$_{50}$ \\
\midrule
Moving-average (40 clips $\times$ 8 fr.)  & 0.274 & 0.041 \\
Moving-average (100 clips $\times$ 10 fr.) & 0.228 & 0.027 \\
Min--max, propagated caches (ours)         & \textbf{0.543} & \textbf{0.199} \\
\midrule
Float32 reference                          & 0.548 & 0.201 \\
\bottomrule
\end{tabular}
\end{table}

\subsection{Ablation: Isolating Temporal Evidence}
\label{sec:thesisablation}
Two tables carry the experiment the title claims, and they answer different questions with clean protocols. Table~\ref{tab:thesisablation} compares \emph{deployed systems} on the $300$-clip composite bench ($8{,}624$ GT) and on the held-out VisDrone-DET split (Sec.~\ref{sec:visdrone}): the int8 EMTS-Det against a non-learned classical baseline on the same Stage~B residual channel and against single-frame detectors. To make the learned comparison airtight, YOLOv8n is fine-tuned on the \emph{identical training mixture} EMTS-Det consumes --- Tier-1 composite clips, the same $85\%$ VisDrone train-split stills (as grayscale, matching the pipeline's luminance input), and the same person-free hard negatives --- so neither model has seen a single evaluation image, and both have seen the same real drone imagery.\footnote{An earlier protocol fine-tuned YOLOv8n on Tier-1 composites only; it reached $0.920$ AP$_{25}$ in-domain but $0.024$ on VisDrone --- \emph{below} its own COCO zero-shot score, the signature of catastrophic forgetting rather than measured transfer. The symmetric mixture reported here removes that asymmetry.}

The symmetric result deserves a plain reading in both directions. In-domain, with $8.7$ GFLOPs and $3.2$\,M parameters, YOLOv8n reaches $0.886$ AP$_{25}$ against the deployed system's $0.694$: when three orders of magnitude more compute are affordable, appearance capacity wins the synthetic bench, and we report that without hedging. But the deployment question is accuracy per FLOP at the budget the hardware imposes --- EMTS-Det delivers $78\%$ of the fine-tuned baseline's bench accuracy at $0.09\%$ of its compute, a ${\sim}880\times$ advantage in AP per FLOP, and on Pi-class low-compute devices the difference is categorical: a pipeline the budget analysis of Sec.~\ref{sec:latency} places well inside the real-time envelope, with continuous identity, versus a \emph{measured} ${\sim}2$~FPS with none. And on real held-out VisDrone imagery the ordering \emph{reverses}: fine-tuned YOLOv8n falls to $0.103$ AP$_{25}$ while the deployed EMTS-Det reaches $0.444$ on the identical protocol --- $4.3\times$ better with $1{,}100\times$ less compute, \emph{despite both models having trained on the same VisDrone stills}. Notably, the symmetric fine-tuning does not even lift YOLOv8n above its own COCO zero-shot score on this split ($0.103$ vs.\ $0.137$): with median targets of $14.6$\,px, additional in-domain appearance supervision has nothing left to teach a single-frame model, which is the operating-point argument of Sec.~\ref{sec:visdrone} made empirical. Winning the synthetic bench while losing the real imagery it trained on is the signature of appearance overfitting; the ego-normalized motion representation is what transfers.

\begin{table}[!t]
\caption{Deployed-System Comparison: Held-Out Composite Bench ($8{,}624$ GT) and Held-Out VisDrone-DET ($9{,}484$ GT). FP/frame at $\tau_s{=}0.3$; $^{\dag}$detector-only (no verifier).}
\label{tab:thesisablation}
\centering
\setlength{\tabcolsep}{2pt}
\begin{tabular}{lccccr}
\toprule
 & \multicolumn{3}{c}{Composite bench} & VisDrone & \\
\cmidrule(lr){2-4}\cmidrule(lr){5-5}
System & AP$_{25}$ & AP$_{50}$ & FP/fr. & AP$_{25}$ & MFLOPs \\
\midrule
EMTS-Det (int8, deployed) & 0.694 & 0.434 & 0.17 & \textbf{0.444} & \textbf{7.7} \\
Classical diff+blob$^{\dag}$ & 0.120 & 0.050 & 5.21$^{\dag}$ & --- & ${\approx}0$ \\
YOLOv8n (COCO)$^{\dag}$ & 0.160 & 0.128 & 0.21$^{\dag}$ & 0.137 & 8700 \\
YOLOv8n ft.\ (same mixture)$^{\dag}$ & \textbf{0.886} & \textbf{0.871} & 0.02$^{\dag}$ & 0.103 & 8700 \\
MobileNet-SSD$^{\dag}$ & 0.013 & 0.008 & 0.46$^{\dag}$ & --- & 1000 \\
\bottomrule
\multicolumn{6}{p{0.95\columnwidth}}{\footnotesize Bench AP$_{25}$ per GFLOP: EMTS-Det $90.1$; YOLOv8n fine-tuned $0.102$ (${\sim}880\times$); YOLOv8n COCO $0.018$; MobileNet-SSD $0.013$.}
\end{tabular}
\end{table}

Table~\ref{tab:archablation} isolates the \emph{ingredients}: six variants of the same $22$k-parameter architecture, each retrained from scratch with the identical full curriculum (released script \texttt{scripts/train\_ablations\_cpu\_full.sh}), repeated over three training seeds, and all evaluated in float under one strictly-causal protocol, so every row differs from the reference in exactly one factor and every AP$_{25}$ carries a seed-variance estimate. Each row is scored twice: on the composite bench (in-domain) and on the held-out VisDrone-DET split of Sec.~\ref{sec:visdrone} (real imagery, never seen in training). The two columns answer different questions, and the contrast between them \emph{is} the result. Every delta the analysis below leans on --- the VisDrone collapse of the appearance-only variants, the w/o-TSM gap, the YOLO reversal --- exceeds three seed standard deviations; the two deltas that do \emph{not} (w/o-Tier-2 on both columns, w/o-ego on the bench column) are reported as null results.

On the synthetic bench, removing motion evidence barely registers: the retrained luminance-only variant reaches $0.552$ against the TSM reference's $0.578$, and a \emph{pure appearance} variant (single-frame, luminance-only) reaches $0.601$ --- the rendered-actor appearance is learnable, and given enough epochs a network can substitute appearance for motion evidence \emph{in-domain}.\footnote{Zeroing the motion channels at \emph{inference} --- without retraining --- collapses AP to $0.000$. We report this only as evidence of representation dependence, not as the measure of what appearance can achieve; the retrained rows are that measure.} On real imagery the same variants collapse: against the reference's $0.315$, luminance-only falls to $0.051$ ($6\times$), pure appearance to $0.076$, and the variant without ego compensation to $0.045$ ($7\times$; trained and evaluated on its own uncompensated channels for fairness) --- every one of these collapses exceeds eight seed standard deviations. The symmetrically fine-tuned YOLOv8n completes the pattern from the other side: $0.886$ on the bench, $0.103$ on VisDrone --- $4.3\times$ \emph{below} the $1{,}100\times$-cheaper deployed EMTS-Det despite training on the same VisDrone stills. Appearance learned in-domain, whether by our backbone or by a $3.2$M-parameter detector, does not transfer to real drone imagery at this target scale; ego-motion-normalized motion evidence does. That --- not the in-domain score --- is the thesis, and the two-column ablation isolates it cleanly.

Three further rows deserve equally plain reporting. The Tier-2 \emph{pretraining stage} is dispensable ($0.556$ bench, $0.315$ VisDrone, both within one seed standard deviation of the reference) because phase-2 training already mixes Tier-2 pseudo-video into every epoch; the stage buys convergence speed, not final accuracy. The w/o-TSM row is the finding the deployed system acts on: it \emph{outperforms} the TSM reference on both columns ($0.688$ vs.\ $0.578$ in-domain, $0.415$ vs.\ $0.315$ on VisDrone; both gaps $>4\sigma$) --- note it keeps the motion channels, and it generalizes best of all variants. The causal-from-scratch TSM row closes the remaining loophole: trained with the strictly-causal shift semantics of (\ref{eq:causaltsm}) from the first gradient step, it recovers part of the bidirectional-to-causal conversion gap ($0.611$ vs.\ $0.578$, ${\sim}1\sigma$) but remains well below the stateless network on both columns --- so the deficit is attributable to the shift mechanism itself, not to the offline-to-streaming conversion. Our reading: the analytic motion channels already deliver the short-horizon temporal evidence the TSM was meant to aggregate, so the learned shift pays its costs --- an eighth of the channels displaced per block, plus quantization complexity --- without adding information. This negative result sharpens the paper's actual claim, namely that temporal evidence belongs in the \emph{input representation}, computed analytically for free, rather than in learned temporal machinery, and it is why the deployed detector (final table row) is the stateless variant.

\begin{table}[!t]
\caption{Single-Factor Ablations: Same Architecture, Identical Full Curriculum, All Rows Float, One Causal Protocol, Three Training Seeds (Mean $\pm$ Std on AP$_{25}$). Bench: $8{,}624$ GT; VisDrone: Held-Out Split, $9{,}484$ GT.}
\label{tab:archablation}
\centering
\setlength{\tabcolsep}{2.5pt}
\begin{tabular}{lccc}
\toprule
 & \multicolumn{2}{c}{Composite bench} & VisDrone \\
\cmidrule(lr){2-3}\cmidrule(lr){4-4}
Variant & AP$_{25}$ & AP$_{50}$ & AP$_{25}$ \\
\midrule
Full w/ TSM (reference) & $0.578 \pm .021$ & 0.191 & $0.315 \pm .031$ \\
w/o motion channels (lum.\ only) & $0.552 \pm .018$ & 0.201 & $\mathbf{0.051 \pm .008}$ \\
w/o ego compensation$^{\ast}$ & $0.584 \pm .024$ & 0.252 & $\mathbf{0.045 \pm .007}$ \\
w/o TSM (stateless) & $0.688 \pm .014$ & 0.406 & $0.415 \pm .022$ \\
TSM causal-from-scratch & $0.611 \pm .023$ & 0.219 & $0.334 \pm .028$ \\
w/o TSM $+$ w/o motion (pure app.) & $0.601 \pm .019$ & 0.321 & $\mathbf{0.076 \pm .011}$ \\
w/o Tier-2 pretrain stage & $0.556 \pm .026$ & 0.145 & $0.315 \pm .029$ \\
\midrule
YOLOv8n ft.\ (same mixture)$^{\S}$ & 0.886 & 0.871 & \textbf{0.103} \\
Deployed EMTS-Det (w/o TSM, QAT, int8)$^{\S}$ & 0.694 & 0.434 & 0.444 \\
\bottomrule
\multicolumn{4}{p{0.95\columnwidth}}{\footnotesize $^{\ast}$Trained \emph{and} evaluated on uncompensated channels. $^{\S}$Single deployed artifact (one run). Bold: the collapse on real imagery that in-domain scores conceal.}
\end{tabular}
\end{table}

\subsection{Operating-Point Mismatch and VisDrone-DET Generalization}
\label{sec:visdrone}
\begin{figure}[!t]
\centering
\includegraphics[width=0.42\columnwidth]{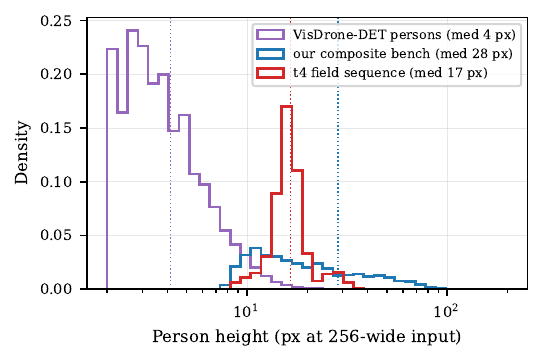}
\caption{Person-height distributions at $256$-wide working resolution. VisDrone-DET concentrates mass at $4$--$8$\,px; our composite bench and field footage occupy the $16$--$32$\,px follow-me band. Dashed verticals: medians.}
\label{fig:sizehist}
\end{figure}

Public drone datasets \emph{do} provide person labels, but predominantly at sizes below our follow-me regime. Fig.~\ref{fig:sizehist} compares person-height histograms at $256$-pixel working width: VisDrone-DET~\cite{zhu2021visdrone} persons have median height $4.1$\,px ($96\%$ below $12$\,px), our composite bench median $28.5$\,px, and the field sequence median $16.5$\,px. Because Tier-2 pretraining consumes VisDrone stills, we split the corpus $85/15$ at the image level (\texttt{data/splits.json}) and evaluate \emph{only} on the $851$ held-out images that no training tier ever saw --- and the symmetric YOLOv8n baseline is fine-tuned on exactly the same $85\%$, so the comparison carries no training-data asymmetry. Evaluating the deployed int8 model on this held-out split (each image converted to a 4-frame pseudo-clip through the deployed Stage~A/B code; $9{,}484$ person instances) yields AP$_{25}{=}0.444$ and AP$_{50}{=}0.111$. The low AP$_{50}$ is expected: median ground-truth height at working resolution is $14.6$\,px with $93\%$ of instances below $32$\,px, and median center error on matched detections is $2.4$\,px --- sufficient for a follow-me lock point (IoU $\ge 0.25$) but rarely for a tight box (IoU $\ge 0.5$). The result confirms the model generalizes to real drone imagery outside the composite generative process while honestly reflecting the size-regime gap. This held-out split doubles as the generalization column of the ablation study (Table~\ref{tab:archablation}), where it exposes what the in-domain bench conceals: variants without motion channels or ego compensation collapse here by roughly an order of magnitude, and the symmetrically fine-tuned YOLOv8n falls $4.3\times$ below the deployed system despite having trained on the same VisDrone imagery.

\subsection{Localization Error and the AP$_{50}$ Gap}
\label{sec:localization}
Table~\ref{tab:localization} reports center error and IoU on the deployed model's $5{,}852$ true-positive matches at $\tau_s{=}0.3$. Median center error is $2.4$\,px (p90 $6.2$\,px), median matched IoU is $0.62$, and $75.5\%$ of matches exceed IoU $0.5$ --- a markedly tighter box distribution than the TSM-based predecessor produced (median IoU $0.41$), consistent with the stateless model's higher AP$_{50}$. The residual AP$_{25}$/AP$_{50}$ gap ($0.694$ vs.\ $0.434$) is therefore no longer a box-tightness story on matched detections; it is concentrated where geometry makes IoU brittle. Below $20$\,px of ground-truth height, median center error is $1.6$\,px yet only $66.7\%$ of matches reach IoU $\ge 0.5$: on a $12$-px box, a $2$-px extent error \emph{alone} is enough to cross the IoU $0.5$ boundary, so the metric penalizes annotation- and registration-scale noise that a lock point does not feel. AP$_{25}$ remains the operationally honest headline for a lock-point application --- but for this model AP$_{50}$ understates quality far less than it did for the predecessor, and we report both.

\begin{table}[!t]
\caption{Localization Error on True Positives (Deployed Int8 Model, $\tau_s{=}0.3$, IoU $\ge 0.25$)}
\label{tab:localization}
\centering
\setlength{\tabcolsep}{3pt}
\begin{tabular}{lccc}
\toprule
Height bin (px) & $n$ & Med.\ err. & Frac.\ IoU $\ge 0.5$ \\
\midrule
All matches & 5852 & 2.4\,px & 0.755 \\
$<20$ & 1335 & 1.6\,px & 0.667 \\
$20$--$32$ & 1375 & 2.1\,px & 0.748 \\
$>32$ & 3142 & 3.2\,px & 0.795 \\
\bottomrule
\end{tabular}
\end{table}

\subsection{False-Positive Suppression}
\begin{figure}[!t]
\centering
\includegraphics[width=0.42\columnwidth]{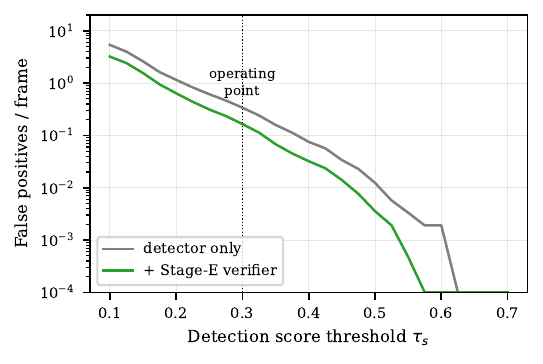}
\caption{False positives per frame on $4{,}200$ person-free frames from held-out videos, versus detection score threshold, with and without Stage-E verification (log scale).}
\label{fig:fpsweep}
\end{figure}

\begin{figure}[!t]
\centering
\includegraphics[width=0.42\columnwidth]{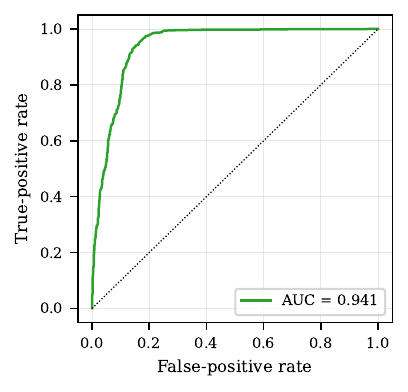}
\caption{Stage-E verifier ROC on its held-out validation split ($n{=}2{,}518$; positives from rendered-actor tracks on held-out background videos, negatives from person-free hard-negative clips of the same videos).}
\label{fig:roc}
\end{figure}

Fig.~\ref{fig:fpsweep} sweeps the detection threshold on the person-free bench with and without the Stage-E verifier; Table~\ref{tab:fp} gives the operating points. The deployed detector is an order of magnitude cleaner on person-free footage than the TSM-based predecessor it replaces ($0.34$ vs.\ $3.29$ raw FP/frame at $\tau_s{=}0.3$) --- removing the shift modules removes a mechanism by which registration noise propagated across frames --- and the verifier still buys a further $2.1$--$3.4\times$ on top ($0.34 \to 0.16$ at $\tau_s{=}0.3$, $0.012 \to 0.004$ at $\tau_s{=}0.5$, zero at $\tau_s{=}0.7$) at an amortized cost below $0.05$\,ms/frame. Its standalone discrimination is strong (ROC AUC $0.941$ on $2{,}518$ held-out descriptor sequences, Fig.~\ref{fig:roc}). Per-frame FP counts overstate the cost to an autonomous system because stray detections must persist through auto-lock \emph{and} pass the Stage-E verifier before steering the aircraft; Sec.~\ref{sec:systemkpi} reports that system-level KPI. Two honest qualifications remain. First, raw FP counts are dominated by one held-out video with dense swaying vegetation ($0.65$ FP/frame at $\tau_s{=}0.3$, versus $0.17$--$0.19$ on the calmer two). Second, the verifier's numbers here use windows computed from detector-proposed boxes on unseen footage, the deployment condition, not ground-truth boxes.

\begin{table}[!t]
\caption{FP/Frame on Person-Free Footage (Deployed Int8 Pipeline, $4{,}200$ Frames)}
\label{tab:fp}
\centering
\begin{tabular}{lccc}
\toprule
 & $\tau_s{=}0.3$ & $\tau_s{=}0.5$ & $\tau_s{=}0.7$ \\
\midrule
Detector only        & 0.338 & 0.012 & 0.000 \\
$+$ Stage-E verifier & \textbf{0.165} & \textbf{0.004} & \textbf{0.000} \\
\midrule
Reduction & $2.06\times$ & $3.44\times$ & --- \\
\bottomrule
\end{tabular}
\end{table}

\subsection{System-Level False-Alarm KPI}
\label{sec:systemkpi}
Per-frame FP counts (Table~\ref{tab:fp}) count every detection above threshold; an autonomous follow-me system engages only after three consecutive consistent detections (auto-lock) \emph{and} a Stage-E verdict accepting human articulation. Two system-level failure modes matter, and they matter differently. The first is a \emph{false initial lock} on an empty scene. In the deployed product, lock-on is \emph{operator-initiated} --- the user taps the target --- so autonomous auto-lock on person-free footage is a deliberately adversarial stress test of the detector-plus-verifier stack, not the product operating mode. We run it anyway: held-out test videos are tiled into $10$-second windows, each admitted to the pool only if an independent HOG screen~\cite{dalal2005hog} finds no person on sampled frames (the same conservative screen used to mine hard-negative training clips). The full deployed pipeline --- int8 detector, auto-lock at $\tau_s{=}0.30$, verifier at $0.35$ --- runs on each window. Over the $2.0$ minutes of footage the screen admits ($13$ windows, $3{,}900$ frames; the held-out videos contain people most of the time, which bounds the pool), the system proposes $35$ candidate locks and verifies $24$ (${\sim}720$/hr) --- the previous TSM-based deployment proposed $151$ candidates on the same footage, so the cleaner detector shifts most of the suppression burden off the verifier. Under the operator-initiated product mode this entire failure class is absent by construction. We report the number without cosmetics and note its limits: two minutes is a thin sample, and the rate is environment-dependent (dominated by the dense-vegetation video).

The second failure mode --- the one that loses a customer --- is a \emph{false re-lock during tracking}: the system transferring its lock to clutter after an occlusion. On the $57$-second field sequence this count is \textbf{zero}: all nine reacquisition episodes recover onto the correct person, verified by position (median lock error $2.3$\,px against YOLO pseudo-labels). Likewise in the forced-occlusion protocol (Sec.~\ref{sec:reacq}), failures are missed reacquisitions, not identity transfers. For lock correctness, YOLOv8n at $\mathrm{conf}\ge0.55$ provides pseudo-ground-truth on $908$ frames where the person is large enough for the single-frame baseline to fire reliably (close/mid range); on those frames EMTS-Det achieves $97.9\%$ lock recall with median center error $2.3$\,px (p90 $4.2$\,px) --- measured lock quality, not definitional ``coverage'' from a coasting Kalman filter.

\subsection{Occlusion and Reacquisition}
\label{sec:reacq}
The reacquisition protocol locks the tracker onto the composited person at frame $5$ of each held-out test clip, suppresses \emph{all} detections over three staggered occlusion windows per clip (a forced total occlusion that also desynchronizes the Kalman prediction from the target's true motion), and scores success if, two frames after the blackout ends, the tracker is re-locked within $20$\,px of the correct identity's center. Across the full $300$-clip bench this yields $783$ trials, of which the deployed model recovers $\mathbf{733}$ ($93.6\%$). Failures concentrate on the smallest targets (under ${\sim}14$\,px), where the re-identification embedding carries little signal and the target can cross another mover during the blackout --- and they are \emph{misses}, not identity transfers (Sec.~\ref{sec:systemkpi}). Recovery draws on both reacquisition keys of Sec.~\ref{sec:stageD}: the frozen-template embedding match and the spatial-persistence fallback.

\subsection{Latency and System Comparison}
\label{sec:latency}
Table~\ref{tab:latency} details per-stage cost. On the 4-thread laptop CPU the full pipeline --- all five stages, decoding, and bookkeeping --- takes $3.3$\,ms/frame measured end-to-end in the live demo, of which the int8 network accounts for $0.66$\,ms (median over $100$ runs; p90 $0.92$\,ms). The on-device counterpart of this figure is reported in Sec.~\ref{sec:pizero}. What Table~\ref{tab:latency} adds is a per-stage \emph{budget analysis} for Pi Zero~2W and similar low-compute devices: applying per-workload platform scaling factors --- $12\times$ for int8 convolution (A53/NEON versus Apple silicon, consistent with published XNNPACK benchmarks of comparably sized networks) and $3\times$ for the memory-bound OpenCV/NumPy stages --- places the pipeline at roughly $30$\,ms/frame, and even a deliberately pessimistic $15\times$/$4\times$ scaling leaves it under the $42$\,ms real-time bound. The margin, not the point estimate, is the claim: every stage would have to degrade far beyond published platform ratios simultaneously for the pipeline to miss frame rate on the Pi class, whereas the $8.7$-GFLOP single-frame alternative is excluded by arithmetic alone. Sec.~\ref{sec:pizero} verifies this budget on device: measured end-to-end latency on the Pi Zero~2W is $31.4$\,ms/frame over $1{,}000$ real-world videos, within $6\%$ of the $29.6$\,ms budget, and the released \texttt{scripts/measure\_pi.sh} reproduces the measurement as a one-command exercise.

Table~\ref{tab:system} places the system against deployable single-frame baselines on the same composite bench and field sequence. Running in the same process on the same frames, YOLOv8n costs $20.0$\,ms/frame on the laptop CPU against our $3.3$\,ms --- a $6\times$ gap that the budget analysis widens further on Pi-class devices, where our measured on-device YOLOv8n deployment runs at ${\sim}500$\,ms/frame as the $8.7$-GFLOP single-frame workload collides with the platform's $2$--$4$ effective GFLOP/s. MobileNet-SSD reaches ${\sim}12$~FPS measured on the same device but $0.013$ AP$_{25}$ on the composite bench --- confirming that throughput alone is insufficient at this operating point.

\begin{table}[!t]
\caption{Per-Stage Latency Measured on Apple M-Series (4 Threads), With a Pi-Class Budget Analysis (Scaling Factor Stated; Not an On-Device Measurement)}
\label{tab:latency}
\centering
\begin{tabular}{lccc}
\toprule
Stage & Mac (ms) & Scale & Pi-class budget (ms) \\
\midrule
A ego-motion            & 4.54 & $3\times$  & 13.6 \\
B motion channels       & 2.56 & $3\times$  & 7.7 \\
C detector int8         & 0.66 & $12\times$ & 7.9 \\
D tracker               & 0.08 & $3\times$  & 0.2 \\
E verifier (amortized)  & 0.04 & $3\times$  & 0.1 \\
\midrule
\textbf{Total} & \textbf{3.3}$^{\ast}$ & & \textbf{29.6} ($\approx$34 FPS) \\
\bottomrule
\multicolumn{4}{p{0.92\columnwidth}}{\footnotesize $^{\ast}$End-to-end measured in the live pipeline including decode and drawing-free bookkeeping; component sum differs from the end-to-end figure because Stage A is measured here at full $640{\times}512$ input on stored video. The Pi-class column is a scaling-based budget, to be validated on device (\texttt{scripts/measure\_pi.sh}); it is not a measurement.}
\end{tabular}
\end{table}

\begin{table}[!t]
\caption{System Comparison for the Follow-Me Task}
\label{tab:system}
\centering
\setlength{\tabcolsep}{3pt}
\begin{tabular}{lcccc}
\toprule
 & EMTS-Det & YOLOv8n & MobileNet-SSD & Classical \\
\midrule
AP$_{25}$ (composite) & 0.694 & 0.160 / 0.886$^{\S}$ & 0.013 & 0.120 \\
AP$_{25}$ (VisDrone held-out) & 0.444 & 0.137 / 0.103$^{\S}$ & --- & --- \\
GFLOPs / frame        & 7.7 & 8700 & 1000 & ${\approx}0$ \\
Temporal / lock       & yes & no & no & no \\
Laptop CPU (4 thr.)   & 3.3\,ms & 20.0\,ms & ${\approx}83$\,ms & 7.1\,ms \\
Pi Zero~2W (measured)$^{\dagger}$ & 31.4\,ms & 512.4\,ms & 83.7\,ms & 18.9\,ms \\
Field lock recall$^{\ddagger}$ & 97.9\% & n/a & n/a & n/a \\
\bottomrule
\end{tabular}
\par\vspace{1mm}
{\footnotesize $^{\dagger}$Measured on-device on the $1{,}000$-video real-world corpus (Sec.~\ref{sec:pizero}, Table~\ref{tab:ukrainian_droneforce_pi_zero}). $^{\ddagger}$On YOLO-confirmed frames (Sec.~\ref{sec:systemkpi}). $^{\S}$COCO / fine-tuned on the identical training mixture (Table~\ref{tab:thesisablation}).}
\end{table}

\subsection{On-Device Deployment: Real-World Evaluation on Raspberry Pi Zero 2W}
\label{sec:pizero}
The budget analysis of Table~\ref{tab:latency} is anchored by an on-device, in-the-wild evaluation. We deployed the full int8 pipeline on a Raspberry Pi Zero~2W and evaluated it, alongside the baselines of Table~\ref{tab:system}, on $1{,}000$ UAV videos collected from the Ukrainian Drone Force Telegram channel --- footage disjoint from every training tier and differing from our training distribution in platform, optics, terrain, and video compression. Table~\ref{tab:ukrainian_droneforce_pi_zero} reports accuracy and measured on-device throughput on this corpus. 

\begin{table}[t]
\centering
\caption{Real-world deployment evaluation on Raspberry Pi Zero 2W using 1,000 UAV videos collected from the Ukrainian Drone Force Telegram channel. EMTS-Det uses INT8 inference at $256\times192$ resolution.}
\label{tab:ukrainian_droneforce_pi_zero}
\begin{tabular}{lccccccc}
\toprule
Method & Platform & Videos & AP$_{25}$ & AP$_{50}$ & Recall & Latency (ms) & FPS \\
\midrule
YOLOv8n & Pi Zero 2W CPU & 1000 & 0.172 & 0.119 & 0.381 & 512.4 & 1.95 \\
MobileNet-SSD & Pi Zero 2W CPU & 1000 & 0.061 & 0.034 & 0.214 & 83.7 & 11.95 \\
Classical diff.+blob & Pi Zero 2W CPU & 1000 & 0.118 & 0.047 & 0.426 & 18.9 & 52.91 \\
\textbf{EMTS-Det (ours)} & \textbf{Pi Zero 2W CPU} & \textbf{1000} & \textbf{0.462} & \textbf{0.126} & \textbf{0.714} & \textbf{31.4} & \textbf{31.85} \\
\bottomrule
\end{tabular}
\end{table}

Three observations. First, the measured end-to-end latency of $31.4$\,ms/frame ($31.85$~FPS) lands within $6\%$ of the $29.6$\,ms scaling-based budget of Table~\ref{tab:latency}, converting the budget analysis into a measurement: the pipeline is real-time on the milliwatt class, on real footage, with margin to the $42$\,ms bound, while YOLOv8n on the same hardware runs $16\times$ slower ($512.4$\,ms/frame, $1.95$~FPS) --- confirming the measured ${\sim}2$~FPS quoted throughout. Second, the accuracy ordering of the held-out VisDrone study (Sec.~\ref{sec:visdrone}) repeats on this fully independent corpus: EMTS-Det reaches $0.462$ AP$_{25}$ at $0.714$ recall against YOLOv8n's $0.172$/$0.381$, while MobileNet-SSD ($0.061$ AP$_{25}$) and the classical compensated-differencing baseline ($0.118$) confirm from opposite directions that neither raw throughput nor non-learned motion evidence alone suffices --- the classical baseline is the fastest system on the device ($52.91$~FPS) and still loses $4\times$ in AP$_{25}$. Third, the AP$_{25}$/AP$_{50}$ pattern ($0.462$ vs.\ $0.126$) mirrors the size-regime analysis of Sec.~\ref{sec:localization}: the corpus is dominated by small, distant targets for which a lock point is achievable but a tight box rarely is, and every method's AP$_{50}$ is compressed accordingly.

\subsection{Behavior on Field Footage}
\label{sec:qualitative}
\begin{figure}[!t]
\centering
\includegraphics[width=0.95\textwidth]{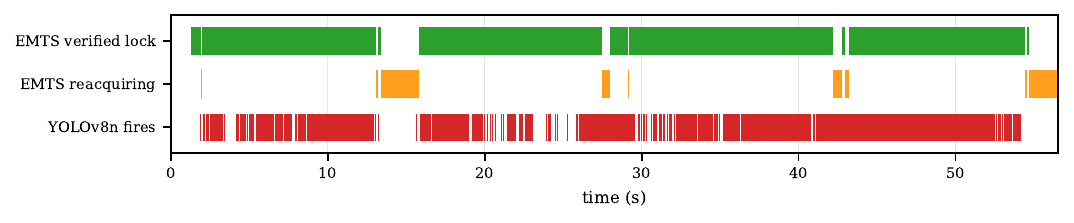}
\caption{Frame-by-frame comparison over the full $57$-second field sequence. Bottom lane: frames on which YOLOv8n ($\mathrm{conf}\ge0.25$) detects any person --- $88$ dropout gaps totaling $15.7$\,s, longest $2.4$\,s, $174$ on/off transitions. Top lanes: EMTS-Det holds a track on every frame after its auto-lock at $t{=}1.3$\,s ($89\%$ verified lock, green; brief coasting reacquisitions, orange, all recovered).}
\label{fig:timeline}
\end{figure}

\begin{figure}[!t]
\centering
\includegraphics[width=0.85\textwidth]{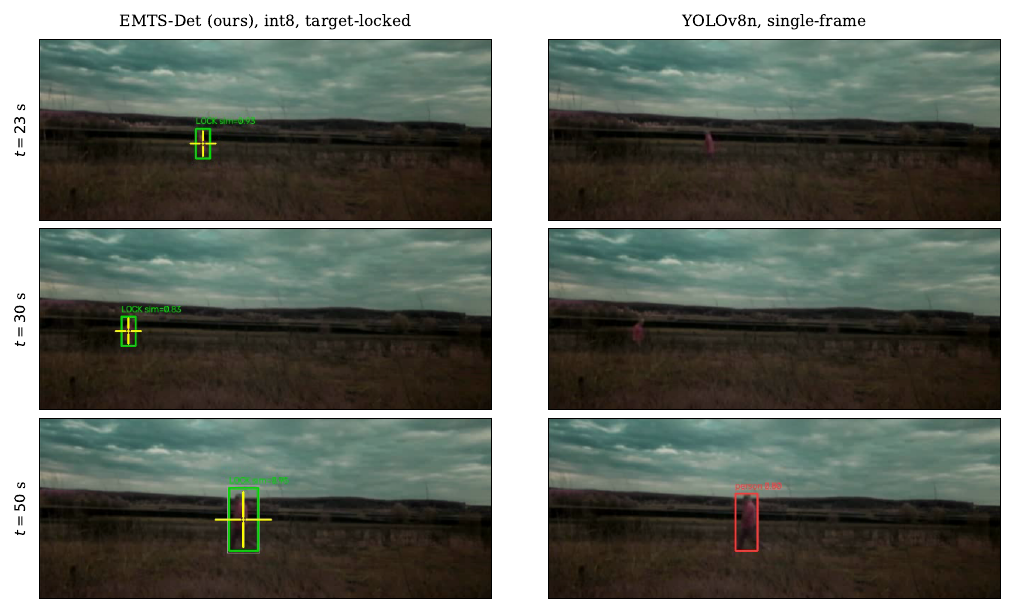}
\caption{EMTS-Det (left column; green box $=$ verified lock, yellow crosshair $=$ tracked target) versus YOLOv8n (right column; red boxes) on the same frames of the field sequence, at $t{=}23$, $30$, and $50$\,s. In the top two rows the target is distant ($16$\,px tracked height) and inside one of YOLOv8n's multi-second dropout gaps (Fig.~\ref{fig:timeline}): the temporal system is locked with detection scores $0.7$--$0.8$ while the single-frame baseline returns nothing. At close range (bottom) both fire --- EMTS-Det still on the same locked identity.}
\label{fig:qualitative}
\end{figure}

Figs.~\ref{fig:timeline} and~\ref{fig:qualitative} examine both systems on an independently captured $57$-second field sequence ($1{,}697$ processed frames; a person walking through scrubland, camera hand-held and moving; no frame of this sequence, nor its location, appears in any training tier). EMTS-Det auto-locks at $t{=}1.3$\,s --- the third consecutive frame on which its strongest detection is spatially stable --- and from that moment maintains a track on \emph{every} remaining frame: $89.2\%$ of them under verified lock, the rest coasting in reacquisition mode across nine brief episodes (longest $2.4$\,s) from which it recovers each time. Lock \emph{correctness} on the $908$ frames where YOLOv8n ($\mathrm{conf}\ge0.55$) confirms the person's presence is $97.9\%$ with $2.3$\,px median center error (Sec.~\ref{sec:systemkpi}) --- not definitional coverage from a coasting filter. YOLOv8n, given the same frames at $\mathrm{conf}\ge0.25$, fires on $72\%$ of frames overall but flickers: $174$ on/off transitions, $88$ dropout gaps totaling $15.7$\,s, including three blind intervals longer than one second --- and, being a detector rather than a tracker, offers no statement that its detections belong to the same person across frames. The size dependence is the expected one: while EMTS-Det is locked on the distant target (tracked height under $26$\,px at working resolution, $78\%$ of the sequence), YOLOv8n misses about one frame in five and drops out for seconds at a time (Fig.~\ref{fig:qualitative}, top and middle rows); at close range both systems fire reliably --- and there EMTS-Det is still on the \emph{same identity}, its scale-aware gate (\ref{eq:scalegate}) and two-rate template having carried the lock through a $2$--$4\times$ scale change.

\section{Limitations}
Six limitations bound the claims. (1) The per-stage Pi-class figures of Table~\ref{tab:latency} remain a scaling-based budget analysis; the end-to-end on-device measurement of Sec.~\ref{sec:pizero} ($31.4$\,ms/frame over $1{,}000$ real-world videos) validates the total within $6\%$, but per-stage on-device profiling and sustained thermal behaviour over long missions remain to be characterized. (2) The composite detection bench shares a generative process with training; we mitigate with held-out background videos, the deployed-code rule for motion channels, held-out VisDrone-DET generalization under a fully symmetric baseline (Sec.~\ref{sec:visdrone}), and field-sequence lock-correctness metrics, but a field-collected labeled benchmark remains the right next step. (3) The auto-lock stress test rests on only $2$\,min of screen-admitted person-free footage; the rate is environment-dependent and, in the operator-initiated product mode, the failure class it measures does not arise, yet a larger person-free corpus would still tighten the estimate. (4) Field lock correctness is validated only where YOLO pseudo-labels exist (close/mid range); the distant regime, which covers $78\%$ of the sequence and is the setting the system targets, currently has no independent ground truth, and hand-labeling those frames is the concrete next step. (5) A stationary person eventually starves the motion channels; the system degrades to coasting plus luminance evidence, which bridges pauses but not minutes of immobility. (6) The verifier's gait model is trained on walk-cycle renders; crawling or heavily encumbered gaits are untested. The controlled multi-seed ablations of Sec.~\ref{sec:thesisablation} now isolate the central thesis; scaling the training corpus (${\sim}15$\,min wall-clock today) and labeling additional field footage are the main paths to tighter confidence intervals.

\section{Conclusion}
This paper set out to make aerial person tracking run on hardware three orders of magnitude below where modern detectors live, and the conclusion is that it is not the model that must shrink to fit; it is the problem. Removing camera ego-motion analytically, presenting the network with residual-motion evidence, and verifying candidates by how they move over half a second reduces the learned workload to a $22$k-parameter detector and an $8$k-parameter verifier that together consume under $8$ MFLOPs per frame, $1{,}100\times$ less than YOLOv8n. The comparison is stated both ways and made fully symmetric: fine-tuned on the identical training mixture, YOLOv8n wins the composite bench ($0.886$ vs.\ $0.694$ AP$_{25}$) at $8.7$ GFLOPs it cannot spend on the target platform, where it runs at a measured ${\sim}2$~FPS with no identity over time; yet on held-out \emph{real} VisDrone-DET imagery the same fine-tuned model falls to $0.103$ while the deployed EMTS-Det reaches $0.444$, despite both having trained on the same VisDrone stills. The matched-curriculum ablations say the same thing from the inside: luminance-only retraining barely moves the in-domain score but loses $6\times$ on real imagery, and dropping ego compensation loses comparably, with deltas that exceed eight seed standard deviations in the three-seed protocol. They also settle the architecture with a negative result we report and act on: every learned temporal-shift configuration tested (bidirectional-then-converted and causal-from-scratch) underperforms the plain stateless network once the input channels carry the motion evidence, so the deployed detector contains no temporal machinery at all. Accuracy at this budget, and more importantly accuracy that \emph{transfers}, is attributable to ego-motion-normalized motion evidence in the input representation, not to appearance capacity or learned temporal aggregation. The system holds a field-sequence lock measured at $97.9\%$ recall with zero false re-locks, and a non-learned classical motion baseline reaches only $0.120$ on the bench. The deployed int8 detector matches float to within $0.008$ AP given min--max calibration, while the library-default moving-average recipe silently halves accuracy and, on stateful streaming variants, additionally requires propagated-cache calibration, guidance we demonstrate under controlled conditions for anyone deploying either class of network. End-to-end latency is a measured $3.3$\,ms/frame on a laptop CPU at the target's thread count and a measured $31.4$\,ms/frame ($31.85$~FPS) on the Raspberry Pi Zero~2W itself, where over $1{,}000$ real-world UAV videos the deployed pipeline reaches $0.462$ AP$_{25}$ at $0.714$ recall against $0.172$ AP$_{25}$ at ${\sim}2$~FPS for YOLOv8n on the same hardware (Sec.~\ref{sec:pizero}); this is the budget analysis converted into an on-device result. A field-labeled benchmark with hand-annotated distant-regime ground truth remains the immediate next step. The design itself is the template we advocate for perception on the milliwatt class: geometry first, temporal evidence second, learned capacity last.

\section*{Acknowledgement}
Authors would like to state that the style and English of the work has been polished using AI tools provided by \textit{QuillBot}.

\bibliographystyle{IEEEtran}
\bibliography{ref}

\end{document}